%% file: main.tex
\pdfoutput=1
\documentclass[11pt]{article}

\usepackage{acl}

\usepackage{times}
\usepackage{latexsym}
\usepackage{enumitem}

\usepackage[T1]{fontenc}
\usepackage[utf8]{inputenc}

\usepackage{microtype}

\usepackage{inconsolata}

\usepackage{xcolor}
\usepackage{tikz-dependency}
\usepackage{booktabs}
\usepackage[export]{adjustbox}
\usepackage{multirow}
\usepackage{bbding}

\title{Parameter-Efficient Instruction Tuning of Large Language Models For Extreme Financial Numeral Labelling}

\author{
    Subhendu Khatuya\textsuperscript{\rm 1} 
    Rajdeep Mukherjee\textsuperscript{\rm 1} 
    Akash Ghosh\textsuperscript{\rm 1} 
    Manjunath Hegde\textsuperscript{\rm 2} \\
    \textbf{Koustuv Dasgupta\textsuperscript{\rm 2} 
    Niloy Ganguly\textsuperscript{\rm 1} 
    Saptarshi Ghosh\textsuperscript{\rm 1} 
    Pawan Goyal\textsuperscript{\rm 1}}\\
    \textsuperscript{\rm 1}Indian Institute of Technology Kharagpur, India\\
    \textsuperscript{\rm 2}Language Modeling, Goldman Sachs\\
    subha.cse143@gmail.com, rajdeep1989@iitkgp.ac.in, akashkgp@gmail.com,\\
    manjunath.y.hegde@gs.com, Koustuv.x.Dasgupta@gs.com, niloy@cse.iitkgp.ac.in,\\ saptarshi@cse.iitkgp.ac.in, pawang@cse.iitkgp.ac.in
}
\newcommand{\dataset}{FNXL}
\newcommand{\modelname}{AttentionXML Pipeline}
\newcommand{\axml}{AttentionXML}
\newcommand{\finer}{FiNER}

\newcommand{\model}{FLAN-FinXC}
\newcommand{\bestmodel}{FLAN-T5-Large with LoRA}

\begin{document}

\maketitle


%\author{
    % Authors
 %   Subhendu Khatuya\textsuperscript{\rm 1},
  %  Rajdeep Mukherjee\textsuperscript{\rm 1},
   % Akash Ghosh\textsuperscript{\rm 1},
   % Manjunath Hegde\textsuperscript{\rm 2},
   % Koustuv Dasgupta\textsuperscript{\rm 2},
   % Niloy Ganguly\textsuperscript{\rm 1},
   % Saptarshi Ghosh\textsuperscript{\rm 1},
   % Pawan Goyal\textsuperscript{\rm 1}
%}
%\affiliations {
    % Affiliations
    %\textsuperscript{\rm 1}IIT Kharagpur, India\\
    %\textsuperscript{\rm 2}Language Modeling, Goldman Sachs\\
    %subha.cse143@gmail.com, rajdeep1989@iitkgp.ac.in, akashkgp@gmail.com, manjunath.y.hegde@gs.com, Koustuv.x.Dasgupta@gs.com, niloy@cse.iitkgp.ac.in, saptarshi@cse.iitkgp.ac.in, pawang@cse.iitkgp.ac.in
%}

\begin{abstract}

% In this work, we study the problem of automatically annotating relevant numerals (GAAP metrics) occurring in the financial statements with their corresponding XBRL tags.
We study the problem of automatically annotating relevant numerals (GAAP metrics) occurring in the financial documents with their corresponding XBRL tags.
%We perform our experiments on the recently released Financial Numeric Extreme Labelling (FNXL) dataset annotated with 2,794 labels.
Different from prior works, we investigate the feasibility of solving this extreme classification problem using a generative paradigm through instruction tuning of Large Language Models (LLMs).
To this end, we leverage metric metadata information to frame our target outputs while proposing a parameter efficient solution for the task using LoRA.
We perform experiments on two recently released financial numeric labeling datasets. 
Our proposed model, \textbf{\model}, achieves new state-of-the-art  performances on both the datasets, outperforming several strong baselines.
%outperforming the strongest baseline with \textbf{39.3\% Macro-F1 gains} and \textbf{17.2\% gains in Hits@1} for FNXL dataset.
We explain the better scores of our proposed model by demonstrating its capability for zero-shot as well as the least frequently occurring tags.
Also, even when we fail to predict the XBRL tags correctly, our generated output has substantial overlap with the ground-truth in majority of the cases.
% W conduct several interesting analyses that show the zero-shot capability of our proposed model, as well as its effective performance on the least frequently occurring labels.

\end{abstract}

\input{Sections/Introduction}

\input{Sections/RelatedWorks}
\input{Sections/ProblemFormulation}
\input{Sections/Methodology}

\input{Sections/Baselines}

\input{Sections/Dataset}

\input{Sections/Experimental_Setup}

\input{Sections/Results}

\input{Sections/Analysis}

\input{Sections/Conclusion}

\input{Sections/Limitations}

\bibliography{custom}

\appendix

% \newpage
\input{Sections/Appendix}
\label{sec:appendix}

\end{document}

%% file: Sections/Introduction.tex
\section{Introduction}

\begin{figure*}[!tbh]
    \centering
     \includegraphics[width=0.95\linewidth]{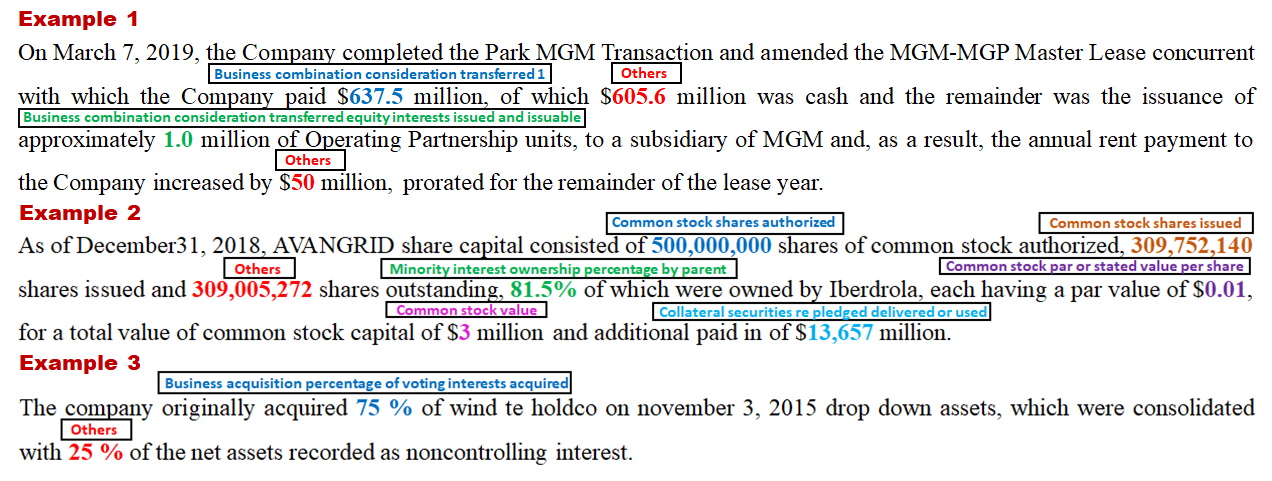}
    \caption{Demonstrating the challenges in the Extreme Financial Numeral Labelling (XFNL) task. Within a financial statement, there are scenarios where every numeral is associated with a distinct XBRL tag, such as in Example 2 (6 distinct tags). Then, there are cases where a mixture of both relevant  and irrelevant numerals (tagged `Others') coexist in the same statement, often within a very limited context, such as in Examples 1 \& 3.} 
    % \pg{Remove Example 3}}
    % This amalgamation of elements further complicates accurate tagging.
    \label{fig:example}

\end{figure*}

The U.S. Securities and Exchange Commission (SEC) mandates publicly traded companies to disclose periodic filings such as quarterly 10-Q \& annual 10-K reports.
These documents are important to finance professionals and investors who rely on SEC filings to make informed investment decisions. 
Each company is directed to follow the \textit{Generally Accepted Accounting Principles} (GAAP) to report the metrics appearing in these documents and tag them using the \textit{eXtensive Business Reporting Language} (XBRL) according to a well-defined taxonomy consisting of thousands of labels.
In a recently released \textbf{FNXL} dataset \cite{sharma2023financial}, such numerals are tagged from a large set of \textbf{2,794 labels}.
Implementing XBRL tagging therefore requires advanced accounting skills to map financial data to the correct XBRL concepts. This requires hiring experts to meticulously review each document and assign appropriate labels which is neither a cost-effective nor a scalable solution. Fig.~\ref{fig:example} shows various challenges involved with the task. 

\begin{table*}[!tbh]
\centering
\small
\resizebox{\linewidth}{!}{
\begin{tabular}{p{0.3\linewidth} | p{0.8\linewidth}}
%\begin{tabular}{c|c}
 \toprule
 Tag & Documentation \\
 \midrule
 \midrule
 common stocks shares issued & Total number of common shares of an entity that have been sold or granted to shareholders (includes common shares that were issued, repurchased and remain in the treasury). These shares represent capital invested by the firm's shareholders and owners, and may be all or only a portion of the number of shares authorized.\\
 % Shares issued include shares outstanding and shares held in the treasury.\\ 
 \midrule
common stocks shares authorized & The maximum number of common shares permitted to be issued by an entity's charter and bylaws. \\ 
 \bottomrule

\end{tabular}
}
\caption{Examples of (XBRL tag, documentation) pairs. We observe that the two tags differ by a single word only whereas their corresponding documentations vary significantly. We take advantage of this distinction while training our \model {} variants. More such examples given in the Appendix A.1}
\label{table:tag_doc}
\end{table*}

\noindent \textbf{Prior works and their limitations:} 
\textit{\finer{}}~\cite{loukas2022finer} formulated the task of identifying relevant numerals from financial texts and labelling them with XBRL tags as an NER problem, and used a
BERT-based sequence labelling approach.
However, the label-set in \finer{} consists of only 139 most frequently occurring XBRL tags.
For practical purposes, a much larger number of XBRL tags/labels are needed to effectively annotate the diverse types of numerals present in these documents. 
More recently, the authors of FNXL dataset~\cite{sharma2023financial}, demonstrated the poor performance of \finer{} when extended to thousands of labels. They also explored an extreme classification methodology, called AttentionXML.
\textit{None of these methods, however, exploit the rich metadata information available with XBRL tags}, to improve the classification performance. Table~\ref{table:tag_doc} provides example of XBRL tag documentations that can help with the labelling task.
Among recent methods that utilize label metadata for better results, \textit{GalaXC}~\cite{saini2021galaxc}, is a GNN-based extreme classification approach that embeds label metadata information in its document-label graph nodes.
\textit{Label Semantics}~\cite{ma2022label} is another generic approach that leverages entity descriptions to solve the standard NER task. However, none of these methods have been applied in the financial domain. We, therefore, adapt these models to use as additional baselines for our task. 

Additionally, all the methods stated above \textit{lack the capacity to identify unseen labels during inference} as they follow a discriminative paradigm.
Generative models, on the other hand, display intrinsic zero-shot capabilities if trained effectively.
In this space, LLMs have achieved impressive performances for a wide range of NLP tasks across several domains~\cite{zhao2023survey}, including finance~\cite{wu2023bloomberggpt, yang2023fingpt}.

\noindent {\bf FLAN-FinXC Framework:} In this work, we show for the first time that generative models (LLMs) can achieve impressive results for the XFNL task. We systematically explore and propose \model{}, a framework of Parameter-Efficient Instruction Tuning for Extreme Classification.% as summarized in Table~\ref{table:flanfinxc_framework}, to gradually improve the performance of LLMs for XFNL. 

Our \model{} framework consists of FLAN-T5~\cite{chung2022scaling} models instruction-tuned with carefully-curated task-specific instructions, as shown in Fig.~\ref{fig:arch}, to generate the appropriate XBRL tag documentations. 
We then make use of an unsupervised \textit{Tag Matcher} module to predict the final XBRL tag for this generated documentation.
We perform extensive experiments to devise a total of five different model variants as part of our proposed \model{} framework, ranging from T5-Base to FLAN-T5-Large, and with varying training strategies.
%where the former has been established, in prior works, to be a better alternative for improving the performance and usability of pre-trained LMs \cite{wei2022finetuned}.
%From Table \ref{table:flanfinxc_framework}, 
We observe that FLAN-T5-Large (instruction-tuned) achieves \textbf{9.4\%} Macro-F1 gains and \textbf{3.5\%} Hits@1 gains, over T5-Large (fine-tuned) for FNXL dataset, both models being architecturally same with 780M parameters. The same trend is also observed in FiNER data.
This highlights the advantages of \textit{Instruction Tuning} similar-sized LLMs over old-fashioned fine-tuning.
Given that training larger models is costly, next we experiment with parameter-efficient (PEFT) techniques, specifically Prefix Tuning~\cite{li2021prefix} and LoRA~\cite{hu2021lora}, to instruction-tune our FLAN-T5-Large models.
%Both techniques require only a small fraction of model parameters to be fine-tuned (0.13\% with Prefix Tuning, and 0.08\% with LoRA), thereby greatly reducing the computation cost and training times. 

%From the second row in Table~\ref{table:flanfinxc_framework}, 
Among these, we observe that FLAN-T5-Large %without PEFT outperforms Prefix Tuning same model.
%Finally, from the last row of Table~\ref{table:flanfinxc_framework}, we find out that training FLAN-T5-Large 
with LoRA outperforms the Prefix-Tuned version with \textbf{2.4\%} Macro-F1 gains and \textbf{5\%} gains in Hits@1, giving the best performance for both the datasets.\footnote{Experimenting with 100 NLP tasks, \cite{ding2023parameter} had also made similar observations that LoRA can sometime even outperform full fine-tuning on certain tasks.}% We also observe the similar trend in our work (refer Table ~\ref{table:result_table}).
%Instruction tuning FLAN-T5-large with LoRA outperforms both the Prefix-Tuned version as well as the model variant instruction-tuned without PEFT with \textbf{11.7\%} and \textbf{17.4\%} Macro-F1 gains, respectively.

%\noindent {\bf Novelty 2: Utilizing XBRL tag documentations:}
%Different from all prior works on XBRL tagging~\cite{loukas2022finer, sharma2023financial}, we leverage the tag metadata for model training by using the elaborate XBRL tag documentations, instead of the relatively shorter tags themselves, to define the targets for our encoder-decoder FLAN-T5 models.
%This decision is driven by our observation that XBRL tags are often very similar to one another, differing only in few words (see examples in Table~\ref{table:tag_doc}), whereas their corresponding documentations allow for a more clear distinction, thereby aiding in the extreme classification task.

Taking advantage of the generative paradigm, parameter-efficient instruction tuning of LLMs, as well as our financial domain-specific novelty (through the use of XBRL tag documentations to improve extreme classification performance), our best model, \textbf{FLAN-T5-Large with LoRA}, outperforms the state-of-the-art \textit{AttentionXML} model with \textbf{39.3\%} Macro-F1 gains and \textbf{17.2\%} Hits@1 gains, thereby achieving new state-of-the-art results for the XFNL task.
We then present several interesting analyses to investigate the reasons for its considerably better performances. 
%Apart from the novelties discussed above, all of which contribute towards obtaining better scores, w
We find that our model achieves impressive zero-shot Macro-F1 scores of \textbf{58.89\%} for the 67 XBRL tags that were unseen during training.
Even for tags that appear fewer than 5 times in the training data, our model is able to achieve 41\% Macro-F1 gains and 23\% Hits@1 gains compared to \textit{AttentionXML}.
Qualitatively, among the instances where we fail to predict the correct XBRL tags, in around 60\% of the cases, our generated tag documentations are very close to the ground truth documentations with Jaccard Similarity scores ranging between 0.6 and 0.85. The proposed model also achieves superior performance (15.22\% Macro-F1 gain and 3.3\% Hits@1 gains) in case of FiNER data containing only most frequent 139 labels. 

% \pg{Report overall improvements.}
%We conducted a detailed analysis of our model's performance taking into account the zero-shot scenarios and labels with the least frequent occurrences.  When dealing with labels appearing between 1 to 10 times, we achieved 50\% accuracy in numerical tagging. With number of examples grows to 50, our Hits@1 score rises to 0.77. For few zero shot labels we get F1-score varies in between 0.60 to 0.80 but for some of them we get poor F1-score of 0.30. We also noticed that absence of explicit instruction template(\textbf{14.2\%} drop) or using tag words (XX\% drop) as target significantly impacts the model performance. As the length of the tag documentation, our target, increases, we observed a marginal drop in performance. However, it's worth noting that the LoRA approach consistently maintains a slight edge over the prefix tuning strategy.
%\noteng{too much detail -may be avoided.}

%\noterm{Till here, the Intro is final from my end.}
%\noteng{There is a confusion which is our method? we should claim one as our paper}
Our contributions can be summarized as follows: 
\begin{itemize}[leftmargin=*]
    \item We propose \model{}, a generative framework, consisting of a suite of instruction-tuned FLAN-T5 models, varying in model sizes and training strategies, to tackle the XFNL task, which is a practically important problem in the finance domain. Different from the prior methods for the task, our models utilize the XBRL tag documentations, instead of considering the tags as just class labels. 
    \item In addition to comparing with the state-of-the-art methods, we adopt several prior works (Label Semantics and GalaXC) for the task, along with devising our own competitive generative baselines (T5-Base and T5-Large). 
    \item For both FiNER as well as FNXL datasets, our best model, \textbf{FLAN-T5-Large with LoRA}, achieves huge improvements over all the baselines. Additional advantages of our best model include substantial performance over rare labels, and its zero-shot capability to tag numerals with labels unseen during training. Our dataset and codes are publicly available at~\url{https://github.com/subhendukhatuya/FLAN-FinXC}.

    %\item  We achieve an impressive Hits@1 score exceeding 90\% for an extensive array of labels. The demonstrated Macro-F1 score of ~\textbf{76\%} highlights its effectiveness in handling our target labels with heavy-tailed distributions. 
    %\item Our model excels at efficiently discriminating irrelevant numerals, which should be appropriately tagged as \textit{`others'}. This proficiency holds true even in scenarios where a complex amalgamation of valid taggable numerals coexist within a very limited financial context. \noteng{is there some numbers?} \pg{Suggest removing this point.}
    %\item Though this work focuses on a particular problem in the financial domain, we believe our experiments bring out key insights into how generative LLMs can be used with parameter-efficient fine tuning approaches for challenging extreme classification problems in any domain, especially where the labels have associated rich semantics. \pg{Moving to conclusion.}
    %\item Our light weight parameter-efficient model is exceptionally well-suited for deployment and for very quick inference. \pg{Are we putting the comparisons for this?} \notesk{Yes, we will put \#of params, model size, etc}\pg{Inference time.}

\end{itemize}

%% file: Sections/RelatedWorks.tex
\section{Related Works}

% \st{In recent years, NLP has found a wide range of applications in the financial domain in areas like risk assessment, regulatory compliance, trading algorithms etc. NLP techniques are used to comb through massive volume of data from different sources and perform tasks like question answering, summarization and sentiment analysis, to name a few.} 
One of the (re)emerging applications of NLP is in the field of Named Entity Recognition (NER), particularly for identifying and categorizing various entities within text.
% Leveraging label semantics and hierarchies can significantly enhance the accuracy and efficiency of NER and related classification tasks.
% . NER aims to identify and categorize the different entities in the text. 
% \st{For standard NLP applications, the categories can be generic like person, location, organization etc. as in %~\cite{sang2003introduction}
% .}
\iffalse{}
The state-of-the-art methods include automated concatenation of embeddings  ~\cite{wang2020automated}, pre-trained contextualized representations using transformers ~\cite{yamada2020luke}, adding external contexts and cooperative learning ~\cite{wang2021improving} and other methods using BERT ~\cite{devlin2019bert} architecture.
In NER, additional context associated with the tags or labels help in improving performance. Label metadata is one such additional information that is often found in financial domain.
\fi
% Metadata like label semantics, label hierarchy can prove useful for NER and classification. 
% ~\cite{ye2021beyond} uses heterogeneous graphs of metadata and labels to perform classification.
% \st{Financial NER can be used for several applications like credit risk assessment %~\cite{alvarado2015domain}.
% } 
The categorization of different entities into labels can also be considered as an extreme classification task~\cite{dahiya2021deepxml}. 
% Extreme classification involves multiclass and multi-label classification on an extremely large label set ~\cite{dahiya2021deepxml}.
% performs extreme multi-label classification by dividing into four sub-tasks to increase efficiency. 
In recent years, XBRL tagging has gained a special importance in financial domain, which involves tagging of numeric values. Datasets created for such a task are the FNXL dataset~\cite{sharma2023financial} and FiNER~\cite{loukas2022finer} which have a very large number of entity types compared to standard NER tasks and thus present challenges to the state-of-the-art NER models.~\cite{sharma2023financial}  reformulated this task as an extreme classification~\cite{you2019attentionxml} problem with a pipeline approach. Financial domain-specific pre-trained language models~\cite{shah2022flue} have been developed, which can be repurposed on financial tasks. Among LLMs pre-trained on financial text (FinLLMs), BloombergGPT~\cite{wu2023bloomberggpt} is a proprietary model that suffers from limited accessibility and lack of transparency in their data collections and training protocols. FinGPT~\cite{yang2023fingpt} is an open-source FinLLM but only suitable for financial sentiment analysis, trading, forecasting and fraud detection tasks.

% In recent years, Large Language Models (LLMs) pre-trained on financial data like 50 billion parameter BloombergGPT ~\cite{wu2023bloomberggpt} are being used for NER tasks as a  benchmark.

% Along with the emergence of generative language models, techniques like prompt tuning and prefix tuning are being explored for a variety of NLP tasks. 
% % Prompt tuning and prefix tuning are parameter-efficient techniques to adapt the Generative Language Models for downstream tasks. Prompt tuning adds a specific prompt to the input instructing the model to perform some tasks. 
% In soft prompt tuning ~\cite{lester2021power}, prompts are trainable and can be learned by backpropagation to improve model performance. Prefix tuning ~\cite{li2021prefix} is inspired from prompt tuning, where the prompt vector is added to each hidden layer of the model and trained. In instruction tuning ~\cite{wei2021finetuned}; ~\cite{chung2022scaling}, the language models are fine-tuned via instructions.

%% file: Sections/ProblemFormulation.tex
\section{Problem Formulation}

\begin{figure}[t]
    \centering
     \includegraphics[width=\linewidth]{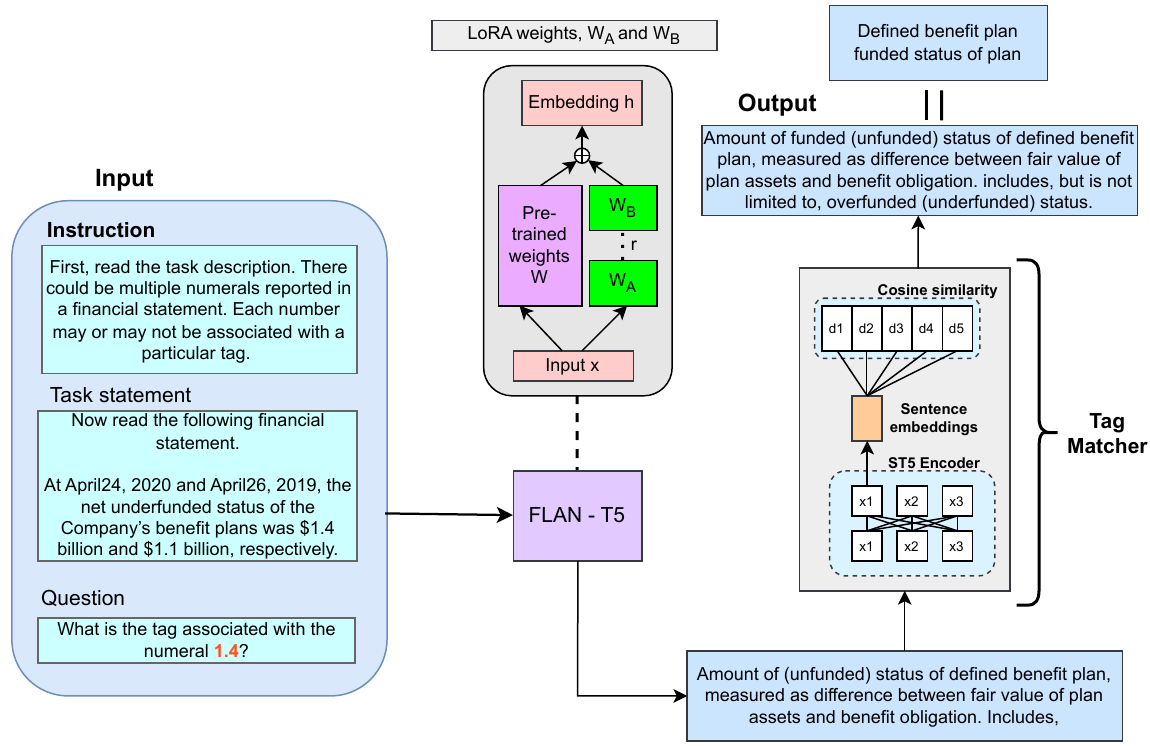}
    \caption{\model{} Architecture. FLAN-T5 takes as   input a task-specific instruction, the financial statement, and a question with a designated target numeral. FLAN-T5-generated tag documentation subsequently flows into the Tag Matcher that predicts the final tag for the given numeral. }

    \label{fig:arch}
\end{figure}

We break the task of XFNL into two stages (with only the first stage requiring supervised training of models), as illustrated in Fig.~\ref{fig:arch}, in order to take advantage of more elaborate XBRL tag documentations. We present a set of diverse annotated examples in Fig.~\ref{fig:example}. Some tag with documentation examples are shown in Table~\ref{table:tag_doc}. 
% In the first stage, we generate the XBRL tag documentation.
% In the second stage, we obtain the final tag from the generated documentation.

In the first stage, we formulate the problem as a generative task using LLMs, where given a financial statement, and a question targeted towards a specific numeral occurring in the statement, the task is to accurately generate its appropriate XBRL \textit{tag documentation} (not the tag) if the numeral is relevant and `Other' if the numeral is irrelevant.
%In order to emphasize the challenging nature of the task, we present a set of diverse annotated examples in Fig.~\ref{fig:example}. Some tag examples are shown in Table \ref{table:tag_doc} to illustrate the utility of tag documentation compared to tag (label).
%\noterm{Not sure if we should write this line here since in Fig.~\ref{fig:example}, annotations are with tags, not tag documentations. If not, Fig.~\ref{fig:example} needs to be referred somewhere in the introduction. I don't think it is referred yet.}
Let $S_i = (w_i^{1},...,w_i^{a},...w_i^{b},..., w_i^{n})$ be the $i^{th}$ statement consisting of $n$ tokens, where $w_i^{a}$ and $w_i^{b}$ be two different numerals with $tag_i^{a}$ and $tag_i^{b}$ being their respective XBRL tag documentations. We prepend an instruction prompt $IP$, containing a natural language description of the task, to the statement $S_i$, as shown in Fig. \ref{fig:arch}.
A question $Q_i^{a}$ is then appended to $S_i$ asking for the tag to be determined for a specific numeral, say $w_i^{a}$.
The modified input $S_i^{a}$ therefore takes the shape $IP || S_i || Q_i^{a}$, where $||$ is a text concatenation operation.
The target answer %$Ans_i^{a}$
$genTag_i^{a} = LLM(S_i^{a})$ for the LLM (FLAN-T5 in our case) therefore becomes: $tag_i^{a}$.

In the second stage, we obtain the final XBRL tag through a separate \textit{Tag Matcher} module, since the entire documentation may not be generated exactly.
Specifically, we obtain the embedding for $genTag_i^{a}$ using a pre-trained state-of-the-art sentence encoder.
We obtain the same for the documentations corresponding to all the available tags.
The one having the highest cosine similarity with the $genTag_i^{a}$ embedding is declared to be the predicted XBRL tag documentation, $predTag_i^{a}$. 
% The corresponding XBRL tag can be easily obtained given the existence of a 1:1 correspondence between XBRL tags and their documentations.

%% file: Sections/Methodology.tex
\section{Methodology}

Our proposed framework \textbf{\model} for the XFNL task is divided into two phases, as depicted in Fig.~\ref{fig:arch}: a supervised generative phase, and an unsupervised documentation-to-tag matching phase.
In the first phase, given a financial statement $S_i$, and a question $Q_i^a$ asking for the XBRL tag (documentation) to be determined for a numeral $w_i^a$ appearing in the sentence $S_i$, we instruction tune FLAN-T5 \cite{chung2022scaling, longpre2023flan} with carefully-curated task-specific instructions (see Fig.~\ref{fig:arch}) to generate the tag documentation $genTag_i^{a}$.
The model is trained to condition on the modified input $S_i^{a}$, as described in the previous section, to generate the target answer one token at a time using auto-regressive decoding. Cross-entropy loss between the generated and true tokens is minimized in the process.
In our case, the entire output generated by the LLM corresponds to the tag description, hence no additional parsing is required.

Our choice for FLAN-T5 is based on the observation that FLAN-T5 models are pre-trained (using instruction tuning) on more than 1.8K tasks, and hence can significantly reduce the amount of fine-tuning steps required if adopted as starting checkpoints for learning new tasks.
Additionally, they achieve strong zero-shot and few-shot performances in comparison to T5~\cite{t5}, their non-instruction-tuned counterpart.
% Also, FLAN-T5 requires lesser steps to converge higher and faster than T5 on single task fine-tuning, therefore offering \textbf{strong green-AI incentives} \cite{longpre2023flan}.

Note that we train FLAN-T5 models to generate the\textit{ XBRL tag documentations} instead of the tag themselves (which is the final target) since the more elaborate documentations allow for a better distinction than their corresponding tags (see Table \ref{table:tag_doc}), thereby aiding in the extremely challenging XFNL task (as demonstrated in Fig.~\ref{fig:example}). Generating the lengthy tag documentations exactly is however difficult, and hence the generated documentation $genTag_i^{a}$ may not exactly match the ground truth.
In the second phase of our proposed framework, we therefore leverage a pre-trained state-of-the-art sentence encoder, as the backbone of our \textit{Tag Matcher} module, as depicted in Fig.~\ref{fig:arch}.
More specifically, we use Sentence-T5-XXL \cite{ni2021sentence}, pre-trained using contrastive loss, to generate 786-dim embeddings for $genTag_i^{a}$ as well as for each of the ground truth tag documentations.
The ground truth tag documentation with the highest cosine similarity (between embeddings) with $genTag_i^{a}$ is considered to be the predicted tag documentation $predTag_i^a$.
Since there exists a 1:1 mapping between XBRL tags and their corresponding documentations, the final predicted tag can be easily obtained from $predTag_i^a$.

In order to investigate the suitability of a generative paradigm for solving the task, we perform a systematic evaluation of several model variants, varying both model sizes and training strategies.

\noindent {\bf Non-FLAN Model Variants:}
First, we compare the fine-tuned performances of T5-Base (220M parameters) with T5-Large (780M parameters). Note that for fine-tuning, given a financial statement $S_i$, the modified input now becomes $S_i^a = S_i || Q_i^a$, where $Q_i^a$ refers to the question targeted towards the numeral $w_i^a$ appearing in $S_i$.
The target (tag documentation) however remains the same.

\noindent \textbf{\model{} Model Variants:}
Next, we compare the instruction-tuned performance of FLAN-T5-Large with fine-tuned T5-Large, both models with identical architectures.
Next, we instruction tune FLAN-T5-Large with PEFT techniques, namely Prefix Tuning and LoRA, respectively requiring only 0.13\% and 0.08\% of model parameters to be updated.
Finally, we instruction-tune FLAN-T5-Large with LoRA to achieve our best results.

%% file: Sections/Baselines.tex
\section{Baselines}

We compare the performance of our proposed models %(including several variations, such as instruction tuning, prefix tuning, LoRA, etc.) 
with the following baselines:

%\noteng{competitive baselines..FINER, AttentionXML}
%\noteng{Modify architecture, GALAXC, MODELS ALREADY TRIED}

%various class of algorithms like named-entity based \finer\ \cite{loukas2021edgar} model, extreme classification based \axml\ and Galaxc \cite{sharma2023financial} model, different base language models like t5, BART and instruction tuning framework FLAN-t5 with various parameter efficient fine tuning (PEFT) frameworks such as LoRA and Prefix-Tuning. 

% \pg{You may also mention which baselines are capable of making use of tag metadata. In Table 3 also, you can have an additional column to indicate this as a tick or cross.} \notesk{will add the column}

\noindent {\bf Baselines already applied for XBRL tagging:}
% We use these two baselines that have already been used for the task. 
\noindent \textbf{(1)~FiNER:}
 ~\cite{loukas2022finer} solved the task as a NER task, but only for top 139 frequent XBRL tags. 
% This is a transformer-based framework with an additional multinomial logistic regression at the last layer  for prediction of the tag.
We adapt this method for our use case (large number of labels).
% They also additionally release SEC-BERT based models which are BERT-BASE models pre-trained on the EDGAR-CORPUS ~\cite{loukas2021edgar}. We showcase the results for six \finer\ based models: three each of BERT-BASE and  SEC-BASE respectively. For each, the three models are {no-masking, [NUM], [SHAPE]}. 
\noindent \textbf{(2)~\axml \space Pipeline:}
% Extreme Classification methods have shown to be effective on real-world datasets where the distribution of data points is extremely skewed and many tail labels often have very few data points to be trained on. 
\cite{sharma2023financial} tackled this problem using a BERT-based sequence-to-sequence tagger.
%using \axml \space framework \cite{you2019attentionxml}. 
% First, they employ a BERT-based sequence-to-sequence tagger to identify relevant or taggable numerals. Then they adapt the \axml \space framework~\cite{you2019attentionxml} to label an entire input instead of labelling a particular span.
 % Here, they fine-tune the bert-base-uncased transformer model, enabling it to extract the contextualized embeddings of subwords and subsequently, a binary logistic regression layer is applied atop these contextual embeddings to accurately predict the relevancy of each numeral. 

%\subsection{Numeral tagger} To tag a particular numeral with a label, we adapt the AttentionXML model. AttentionXML attempts to label an entire input instead of labelling a particular span.
% We adapt the AttentionXML model and through variations in input focus on a particular span. The module takes as input a particular sentence with the target numeral marked in a special way. We make the following changes to AttentionXML:
% 1) We replace the Glove + BiLSTM layer of the Attention-Aware Deep Model with BERT representations of the input. 
% 2) We apply a local attention layer after obtaining the BERT representations. The local attention layer attends to the span of the target numeral.
% 3) We also experiment with the concept introduced by \finer to avoiding defragmentation and replace numerals with the [NUM], [SHAPE] token. Here, the target numeral is left untouched and the others are replaced by the token. We showcase our masking strategies in Figure ~\ref{fig:token_replacement_eg}.

\noindent {\bf New baselines that we adopt for XBRL tagging:}
Additionally, we adopt two diverse methods for the task for XBRL tagging, which can utilize the label semantics of XBRL tags.
\noindent \textbf{(3)~GalaXC:}
~\cite{saini2021galaxc} applies collaborative learning over document-label graphs that allows additional information like label metadata to be incorporated. 
% GalaXC utilizes a GNN architecture and employs a joint graph over documents and labels and learns node representations using graph convolutions. The learning of document and label embeddings is aided with label metadata. 
% A novel attention scheme is also proposed where multiple representations are learned from each graph node. These are termed as the multi-resolution representations for documents and labels.
% High-capacity extreme classifiers and a label-wise attention mechanism are used over the embeddings, and a scalable mechanism is used to incorporate test documents into the graph.
\textbf{(4)~Label Semantics}~\cite{ma2022label} leverages the entity description to solve a standard NER task. Lastly, with the emergence of ChatGPT\cite{openai2022chatgpt}, we were curious to check ChatGPT’s performance for this task for 500 random samples using \textit{gpt-3.5-turbo} \footnote{\url{https://platform.openai.com/docs/models/gpt-3-5}} API.

% A neural architecture consisting of two BERT encoders is used, where one encoder is used to encode the document and another one to encode the labels. 

% The model uses two encoders and is trained to align the named entity representations obtained from the first encoder with the label representations obtained from the second encoder into a common embedding space. We adapt this method for our use-case where the tag words are used as label metadata.

% \subsection{Prompt Tuning}
% Prompt tuning is a parameter-efficient fine-tuning technique where an input prompt is added to condition the model on the output generation. 

%% file: Sections/Dataset.tex
\section{Dataset  \& Evaluation Metrics}

\noindent {\bf Dataset:} We perform our experiments on the recently released \dataset\space dataset~\cite{sharma2023financial} containing 10-K documents\footnote{\url{https://tinyurl.com/t43mwd5m}}
for 2,339 companies and FiNER \cite{loukas2022finer} dataset having most frequent (at least 1000 appearances) 139 labels. FNXL dataset contains a total of 79,088 sentences containing 142,922 annotated numerals, with a tag set of \textit{2,794 distinct tags} that follows a heavy tail distribution. Further, the test set contains \textit{67 XBRL tags that are not part of the train or validation sets}.
% Figure~\ref{fig:example} shows a few examples of sentences from this dataset, along with the XBRL tags for the numerals contained in the sentences.
Fig.~\ref{fig:example} shows a few annotated examples from this dataset.
Each tag is associated with a pre-defined textual description known as its `tag documentation' (see Table~\ref{table:tag_doc}).
% Table~\ref{table:tag_doc} shows some examples of XBRL tags and their documentations. 
The tag documentations have an average length of 28 words and a maximum length of 273 words.
% Note that, while certain tag-pairs may exhibit subtle distinctions, their accompanying documentations vary significantly (as shown in Table~\ref{table:tag_doc}). 
%Domain experts leverage the information provided in the documents during manual annotation. Our proposed model is also designed to utilize these documentations. 

% \noindent {\bf Train-test split:} 
% The dataset was split in approx. 80\%-8\%-12\% train-validation-test splits by~\citet{sharma2023financial}. We use the same splits to train and evaluate the models. 
% To check if our model possesses zero-shot capability, we deliberately included 67 labels in the test set that were \textit{not} present in the training data. 

\noindent {\bf Metrics:}
To ensure a comprehensive evaluation of all models, we employ the following metrics: 1)~Macro-Precision, 2)~Macro-Recall, 3)~Macro-F1, 4)~Hits@1. 
% The macro-averaged F1 score is determined through the arithmetic mean of per-class F1 scores. 
In financial numeral labelling, where equal importance is given to all tags, macro based metrics (precision, recall, F1) are a good choice as it treats all classes equally, irrespective of their frequency. We report \textit{Hits@1} metric to showcase the usability of this system from business perspective, recommending the \textit{top tag} to domain experts.

% We identify that a major business use-case of this system is to recommend the \textit{top tag} to subject matter experts (SME) for a particular numeral which she may use to quickly produce the correct annotation. Thus, we report \textit{Hits@1} metric to showcase the usability of this system from business perspective.

% In our \dataset\ dataset, we see that top 100 frequently occurring labels (each containing more than 100 data points) out of 2794 correspond to 58.79\% of our total data points and least 1856 frequently occurring labels (each containing less than 20 data points) constitute 8.34\% of our total data points. 

 % The sentences have an average length of 197 characters and a maximum length of 3074. 
 % For more details of the data statistics, refer to  Table ~\ref{table:data_stats}. Please refer to Appendix ~\ref{sec:appendix} for the frequency distribution of the labels.

%% file: Sections/Experimental_Setup.tex
\section{Experimental Setup}

% For all our model variants, namely T5-Base\footnote{https://huggingface.co/t5-base} (220M), T5-Large\footnote{https://huggingface.co/t5-large} (780M), FLAN-T5-Large\footnote{https://huggingface.co/google/flan-t5-large} (780M), and FLAN-T5-XL\footnote{https://huggingface.co/google/flan-t5-xl} (3B), we obtain the pre-trained checkpoints from the \textit{Huggingface} Library.
For all our model variants (experiments performed on Tesla V100 32GB GPUs),
% , namely, T5-Base, T5-Large, and FLAN-T5-Large,
we obtain the pre-trained checkpoints from the \textit{Huggingface} Library\footnote{\url{https://huggingface.co/}}.
For instruction tuning FLAN-T5-Large with Prefix Tuning, the \textit{prefix length} was set to 20.
For training the models with LoRA, the \textit{rank} for the trainable decomposition matrices was set to 2. FLAN-T5-Large models were instruction-tuned for 10 epochs with a learning rate (lr) of $1e-4$ (training time: 1hr 22 minutes/epoch, inference time: 2 minutes/sample); with an lr of $1e-2$ with Prefix Tuning (training time: 29 minutes/epoch, inference time: 2 minutes/sample); and with an lr of $5e-4$ with LoRA (training time: 56 minutes/epoch, inference time: 2 minutes/sample).
% FLAN-T5-XL was trained with LoRA for 5 epochs with an lr of $5e-4$ (training time:  2hr 20 minutes/epoch, inference time: 6 minutes/sample).
These hyperparameters were selected based on the best Macro-F1 results on the validation set.
% For all our experiments, the maximum input length was set to 128 and the maximum length of output to be generated was set to 30. 
% In all experiments, the input length was limited to 128, and the output to be generated was set to 30. 
% All our experiments are performed on Tesla V100 32G GPUs.

% Prefix Tuning and LoRA respectively require 0.13\% and 0.08\% of model parameters to be fine-tuned, thereby greatly reducing the computation cost and training times.

Note that the tag information is \textit{not} included in the input prompt. Rather, it is generated during the decoding phase based on the question asked. The input prompt containing the question gets encoded and it is much smaller than the maximum input context length. Hence, there is no truncation during encoding. The average tag information length is 28 tokens and we set the length of output to be generated to 30 tokens. Hence, in a few cases, truncation may occur during the decoding phase. However, the crucial tag information is typically present in the initial tokens, and our tag matcher ensures correct labels by comparing generated tags with the descriptions for all available tags in the dataset. Thus any potential truncation in a few cases does not pose any risk to our model's performance.

\noindent {\bf Instruction Selection:} 
While experimenting, we gradually improved our instructions. First, we tried out a prompt without any task-specific instruction, thereby consisting of the input financial statement and the question for a given numeral. Then, we included a simple task description and repeated our experiments. Finally, we made the task description/instruction more elaborate which led us to obtain the best results. On the target side, we tried out two variants, generating labels vs. generating the label descriptions. The latter resulted in significant performance improvement (refer Table~\ref{table:ablation_study}). This iterative process of instruction selection was guided by the performance on the validation set.

%% file: Sections/Results.tex
\begin{table*}
  \centering
    \resizebox{\linewidth}{!}{
      \begin{tabular}{l|c|c|c|c|c|c|c|c}
        \toprule
        Model & \multicolumn{8}{c}{Dataset} \\
        \midrule
        & \multicolumn{4}{c|}{FNXL \cite{sharma2023financial}} & \multicolumn{4}{c}{FiNER \cite{loukas2022finer}} \\
        \cmidrule{2-9}
        & M-P & M-R & M-F1 & Hits@1 & M-P & M-R & M-F1 & Hits@1 \\
        \midrule
        FiNER (bert-base) & 49.17 & \underline{49.71} & 47.13 & 75.34 & 72.60 & 81.10 & 76.61 & 81.50\\
         \finer{} (sec-base) & 47.76	& 48.87 & 46.20 & 74.67 & 81.11 & \underline{83.20} & 82.14 & 82.30\\
         Label Semantics & 46.35 & 45.12 & 45.72 & 71.25 & 71.50 & 80.15 & 75.57 & 80.25\\
         GalaXC & 46.91 & 44.81  & 45.81 & 72.97 & 72.20 & 80.95 & 76.32 & 81.10 \\
        \modelname & \underline{50.69} & 48.51 & \underline{47.54} & \underline{76.76} & \underline{82.15} & 82.30 & \underline{82.22} & \underline{83.25}\\
        \midrule
        ChatGPT (500 samples) & 11.13 & 7.68 & 9.08 & 19.6 & 20.12 & 15.67 & 17.61 & 22.35\\
        T5-Base & 59.94 & 49.48 & 54.21 & 79.21 & 86.92 & 84.35 & 85.61 & 83.45\\
        T5-Large & 61.87 & 58.46 & 60.11 & 83.26 & 88.12 & 85.10 & 86.58 & 84.12\\
        \midrule
        
        %\multicolumn{5}{c}{\model{} Model Variants} \\
        
        %\midrule
        FLAN-T5-Large & \textbf{66.21} & 65.34 & 65.77 & 86.21 &\textbf{ 92.10} & 96.35 & 94.17 & 85.89\\        
        % Prefix-Tuning (T5-Large) & 65.65 & 63.48 & 64.54 & 81.81 \\
          FLAN-T5-Large with Prefix-Tuning & 65.10 & 64.21 & 64.65 & 85.69 & 90.18 & 94.35 & 92.21 & 85.12\\        
         % \quad w/ LORA & 71.77 & \textbf{74.74} & 73.22 & 89.98 & -- & -- & -- & --\\
         FLAN-T5-Large with LORA & 65.14 & \textbf{67.36} & \textbf{66.23} & \textbf{89.98} & 91.84 & \textbf{97.85} & \textbf{94.74} & \textbf{86.03}\\

        \bottomrule
      \end{tabular} 
      }
   \caption{Performance evaluation based on Macro \& Hits@1 metrics for FNXL dataset and FINER dataset. The best performance is highlighted in bold, and the strongest baseline result is underlined.}
   \label{table:result_table}
\end{table*}

\section{Main Results}

We report the results of our proposed model variants and various baselines in Table~\ref{table:result_table} for both FNXL and FiNER dataset. 
Among the baselines, \finer\ does not perform well for large number of entity labels. On the other hand, while the `Label Semantics' method leverages tag words within an NER framework, its scalability is constrained when dealing with a large number of entity labels, ultimately leading to poor performance. 
The \axml\ pipeline performs  better than all other baselines. 
GalaXC and `Label Semantics' have very similar performance, with GalaXC giving slight edge.
ChatGPT's performance for this complex task is not satisfactory.

We also demonstrate the advantage of training bigger models for extreme financial numeral labelling for FNXL dataset, by comparing the results of T5-Large (780M parameters, Macro-F1 60.11) with T5-Base (220M parameters, Macro-F1 54.21). 
Note that these baselines that we devise for the task, already outperform the existing state-of-the-art.
%achieving \textbf{5\%} Macro-F1 gains and \textbf{2.8\%} Hits@1 gains. 

Next, we turn to different variations of our FLAN-FinXC framework (listed in the lower part of Table~\ref{table:result_table}). 
First, we demonstrate the advantage of instruction tuning over fine-tuning by the fact that FLAN-T5-Large (Macro-F1 65.77) substantially outperforms T5-Large (Macro-F1 60.11) for FNXL dataset.
%achieving \textbf{9.4\%} Macro-F1 gains and \textbf{5.6\%} Hits@1 gains.
Given that training larger models is costly, %and makes the comparisons with our baselines unfair, 
next we experiment with parameter-efficient (PEFT) techniques, specifically Prefix Tuning~\cite{li2021prefix} and LoRA \cite{hu2021lora}, to instruction tune our FLAN-T5-Large models.
Both techniques require only a small fraction of model parameters to be fine-tuned (0.13\% with Prefix Tuning, and 0.08\% with LoRA), thereby greatly reducing the computation cost and training times. 
Instruction tuning FLAN-T5-large with LoRA (Macro-F1 66.23) outperforms Prefix-Tuned version of the same model (Macro-F1 64.65) and has a slight edge over the model variant instruction-tuned without PEFT (Macro-F1 65.77). 
%with \textbf{11.7\%} and \textbf{17.4\%} Macro-F1 gains, respectively.
Our best results are obtained by instruction tuning FLAN-T5-Large with LoRA (only 0.08\% of 780M parameters to be finetuned). This model outperforms the state-of-the-art \textit{AttentionXML} model with \textbf{39.3\%} Macro-F1 gains and \textbf{17.2\%} Hits@1 gains, thereby achieving new state-of-the-art results for FNXL dataset. For FiNER dataset, we achieve 15.22\% Macro-F1 gain and 3.3\% Hits@1 gains.

\if{0}
% old table
\begin{table}[t]
  \centering
    \resizebox{\linewidth}{!}{
      \begin{tabular}{l|l|c|c|c|c}
        \toprule
        Model & Dataset & M-P & M-R & M-F1 & Hits@1 \\
        \midrule
        FiNER (bert-base) &  & 49.17 & \underline{49.71} & 47.13 & 75.34 \\
         \finer{} (sec-base) & & 47.76	& 48.87 & 46.20 & 74.67 \\
         Label Semantics & FNXL\cite{sharma2023financial} & 46.35 & 45.12 & 45.72 & 71.25 \\
         GalaXC & & 46.91 & 44.81  & 45.81 & 72.97 \\
        \modelname & & \underline{50.69} & 48.51 & \underline{47.54} & \underline{76.76} \\
        \midrule        
        T5-Base & & 59.94 & 49.48 & 54.21 & 75.01 \\
        T5-Large & & 61.87 & 52.76 & 56.95 & 77.16 \\
        \midrule
        
        %\multicolumn{5}{c}{\model{} Model Variants} \\
        
        %\midrule
        FLAN-T5-Large & & 63.11 & 61.62 & 62.35 & 79.25 \\        
        % Prefix-Tuning (T5-Large) & 65.65 & 63.48 & 64.54 & 81.81 \\
          \quad w/ Prefix-Tuning & & 71.81 & 60.30 & 65.55 & 83.69 \\        
         \quad w/ LORA & & 71.77 & \textbf{74.74} & 73.22 & 89.98 \\
         FLAN-T5-XL w/ LORA & &\textbf{80.31} & 71.73 & \textbf{75.77} & \textbf{90.84} \\
        \midrule
        
        \finer{} (sec-base) & & 81.11 & 83.20 & 82.14 & 82.30 \\        
        % Prefix-Tuning (T5-Large) & 65.65 & 63.48 & 64.54 & 81.81 \\

         \modelname & FiNER \cite{loukas2022finer} & 82.15 & 84.32 & 83.22 & 83.25 \\ 
        FLAN-T5-Large w/ LORA & & \textbf{91.84} & \textbf{97.85} & \textbf{94.74} & \textbf{86.03} \\
        \bottomrule
      \end{tabular} 
      }
   \caption{Performance evaluation based on Macro \& Hits@1 metrics for FNXL dataset and FINER dataset. The best performance is highlighted in bold, and the strongest baseline result is underlined.}
   \label{table:result_table}
   % 65.14	67.36	66.23	89.98
\end{table}
\fi

\if{0}
% old table
\begin{table}
  \centering
    % \resizebox{\linewidth}{!}{
      \begin{tabular}{l|c|c|c|c|c|c|c|c}
        \toprule
        Model & \multicolumn{8}{c}{Dataset} \\
        \midrule
        & \multicolumn{4}{c|}{FNXL\cite{sharma2023financial}} & \multicolumn{4}{c}{FiNER \cite{loukas2022finer}} \\
        \cmidrule{2-9}
        & M-P & M-R & M-F1 & Hits@1 & M-P & M-R & M-F1 & Hits@1 \\
        \midrule
        FiNER (bert-base) & 49.17 & \underline{49.71} & 47.13 & 75.34 & -- & -- & -- & --\\
         \finer{} (sec-base) & 47.76	& 48.87 & 46.20 & 74.67 & 81.11 & 83.20 & 82.14 & 82.30\\
         Label Semantics & 46.35 & 45.12 & 45.72 & 71.25 & -- & -- & -- & --\\
         GalaXC & 46.91 & 44.81  & 45.81 & 72.97 & -- & -- & -- & -- \\
        \modelname & \underline{50.69} & 48.51 & \underline{47.54} & \underline{76.76} & 82.15 & 84.32 & 83.22 & 83.25\\
        \midrule        
        T5-Base & 59.94 & 49.48 & 54.21 & 75.01 & -- & -- & -- & --\\
        T5-Large & 61.87 & 52.76 & 56.95 & 77.16 & -- & -- & -- & --\\
        \midrule
        
        %\multicolumn{5}{c}{\model{} Model Variants} \\
        
        %\midrule
        FLAN-T5-Large & 63.11 & 61.62 & 62.35 & 79.25 & -- & -- & -- & --\\        
        % Prefix-Tuning (T5-Large) & 65.65 & 63.48 & 64.54 & 81.81 \\
          \quad w/ Prefix-Tuning & 71.81 & 60.30 & 65.55 & 83.69 & -- & -- & -- & --\\        
         \quad w/ LORA & 71.77 & \textbf{74.74} & 73.22 & 89.98 & -- & -- & -- & --\\
         FLAN-T5-Large w/ LORA & \textbf{80.31} & 71.73 & \textbf{75.77} & \textbf{90.84} & \textbf{91.84} & \textbf{97.85} & \textbf{94.74} & \textbf{86.03}\\

        %\finer{} (sec-base) & & 81.11 & 83.20 & 82.14 & 82.30 \\        
        % Prefix-Tuning (T5-Large) & 65.65 & 63.48 & 64.54 & 81.81 \\

         %\modelname & FiNER \cite{loukas2022finer} & 82.15 & 84.32 & 83.22 & 83.25 \\ 
        %FLAN-T5-Large w/ LORA & & \textbf{91.84} & \textbf{97.85} & \textbf{94.74} & \textbf{86.03} \\
        \bottomrule
      \end{tabular} 
      % }
   \caption{Performance evaluation based on Macro \& Hits@1 metrics for FNXL dataset and FINER dataset. The best performance is highlighted in bold, and the strongest baseline result is underlined.}
   \label{table:result_table_old}
\end{table}
\fi

%\fi

% \begin{table}
% \centering
% \small 
% \begin{tabular}{c|c|c}
% \toprule
% Tag Matcher & FLAN T5-Large & FLAN T5-XL \\
% \midrule
% S-BERT-L6 & 69.78 & 72.71 \\
% S-BERT-L12 & 70.18 & 73.31 \\
% T5-xxl & 73.22 & 75.77 \\
% \bottomrule
% \end{tabular}
% \caption{Performance of different sentence embedding methods. Among them, the F1-score of T5-XXL stands out as the most impressive for both variants of our model.}
% \label{table:perf_tag_matcher}
% \notesk{Remove Table 4, check where it is referred}
% \end{table}

% \notesk{Add two hard examples to showcase that our method wins over classification framework}

%% file: Sections/Analysis.tex
\section{Analysis}
\label{sec:analysis}

We now present different analyses of our best model (FLAN-T5-Large with LoRA), and the closest baseline (\axml{} Pipeline). 
The following analysis is specific to the FNXL dataset, chosen for its more extensive range of labels compared to other datasets, emphasizing its practical relevance.

% \subsection{Analysis}\label{sec:bucket_analysis}

\subsection{Performance on least frequent labels}

% \noindent \textbf{Performance on rare (least frequently occurring) labels:}
% In XBRL tagging, correct tagging of rare labels is essential for reliable financial reporting, since the rare labels often represent significant regulatory requirements. However, they pose challenges due to imbalanced data, limited training instances, and sparse data distribution. 

Accurate tagging of infrequent labels in XBRL is crucial for reliable financial reporting, yet it presents challenges stemming from imbalanced data, scarce training instances, and sparse data distribution. We observe our \model{} performs substantially better than \axml{} for the rare labels. To demonstrate this, we group the tags into various buckets based on their frequency of occurrence in the training set.

\begin{figure}[!tbh]
\centering	
{
\includegraphics[height=4.5cm]{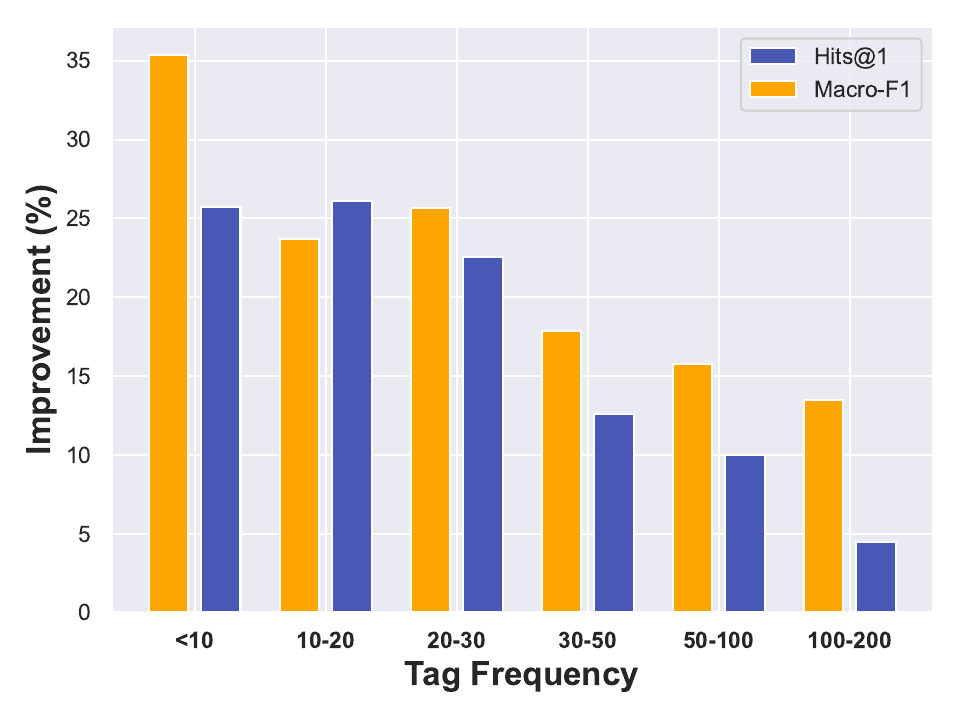}
}
\caption{Relative improvement in performance achieved by FLAN-T5-Large with LoRA over \axml{}  Pipeline, for the least frequent labels under various frequency buckets}
\label{fig:least_frequent_per}
\end{figure}

Fig.~\ref{fig:least_frequent_per} shows the percentage improvement in Hits@1 and Macro-F1 that is achieved by \model{} over \axml{} Pipeline for the tags in every bucket. 
%performance of our best model for the least frequency tags, where we have grouped the tags into distinct buckets according to their frequency of occurrence in the training data. 
We see that our model can effectively identify and tag rare financial concepts. 
%We've grouped tags into distinct buckets according to their frequency of occurrence in the training data, and we're evaluating the performance of each bucket. 
%Figure~\ref{fig:least_frequent_per}, shows the relative improvement of performance metrics across different frequency ranges. 
Notably, even for tags that appear fewer than 10 times in the training data, our model is able to achieve 35.3\% improvement in Macro-F1 score and 25\% in Hits@1 over the closest baseline.

\subsection{Zero-Shot Capability}
One of the key strengths of our proposed model is its zero-shot capability, i.e., its ability to generate tags/labels for which it has \textit{not} been explicitly trained. 
While SOTA discriminative models in this domain often require specific fine-tuning or retraining to handle new tags, our generative model transcends these limitations. 
%This adaptability not only saves time and resources spent on retraining, but also enables rapid deployment of the model in real-world situations. 

\begin{table}[hbt!]
\centering
\small 
\begin{tabular}{p{0.7\columnwidth}|c}
\toprule
Tag & F1-score \\ 
\hline
foreign currency transaction gain before tax & 0.85 \\

%\midrule
commercial paper at carrying value & 0.85 \\
accounts payable other current & 0.80\\ 
%\midrule
recognition of deferred revenue & 0.76 \\
%\midrule
 available for sale debt securities gross unrealized gain & 0.66 \\
%\midrule
%payments to acquire held for sale real estate & 0.26 \\
\bottomrule
\end{tabular}
\caption{Zero-Shot Performance of \bestmodel{}, for a few XBRL tags absent in the training set of FNXL dataset.}
\label{table:zero-shot-examples}
\end{table}

Recall that the test set included 67 new labels that were not present in the train set. 
We observe that our best model achieves a Macro-F1 of \textbf{58.89} over these 67 unseen labels, which is a commendable performance. 
Table~\ref{table:zero-shot-examples} shows the performance (F1-score) of our best model for few such tags.

\begin{table}[!thb]
  \centering
    \resizebox{\linewidth}{!}{
      \begin{tabular}{l|c|c}
        \toprule
        Model & Macro-F1 & Hits@1 \\
        \midrule

        FLAN-T5-Large with LoRA & 66.23 & 89.98 \\
        \quad w/ S-BERT-L12 as Tag Matcher & 63.11 & 88.13 \\
        \quad w/ S-BERT-L6 as Tag Matcher & 62.87 & 87.72\\
        \quad w/o instruction prompt & 56.46 & 76.55 \\
        \quad w/o tag metadata & 53.12 & 73.14 \\

        \bottomrule
      \end{tabular} 
      }
   \caption{Results for our ablation studies of FNXL dataset}
   \label{table:ablation_study}
\end{table}

\subsection{Ablation Study}
% \noindent \textbf{Ablation Study:}
We now try out various ablations over our best model to understand the significance of different modules.
First, as ablations of our \textit{Tag Matcher} module, we replace Sentence-T5-XXL (used in our best model) with Sentence-BERT~\cite{reimers2019sentence} to generate the embeddings for the XBRL tag documentations. We experiment with two versions of Sentence-BERT, one with 6 encoder layers and other with 12 encoder layers, and report our findings in Table~\ref{table:ablation_study}. 
We observe a drop in performance (compared to our best model \model{} with LoRA) in both cases, thereby showing the better efficacy of Sentence-T5-XXL over Sentence-BERT as sentence encoder.

Next, we instruction-tune FLAN-T5-Large (with LoRA) \textit{without the instruction prompt} containing specific instructions for the XFNL task.
Given a financial statement $S_i$, the modified input now becomes $S_i^a = S_i || Q_i^a$, where $Q_i^a$ refers to the question targeted towards the numeral $w_i^a$ appearing in $S_i$, and $||$ is a text concatenation operation.
From the fourth row in Table~~\ref{table:ablation_study}, we observe a huge drop in performance (e.g., Macro-F1 drops from 66.23 to 56.46), thereby demonstrating the importance of aligning FLAN-T5 fine-tuning with task-specific instructions to tackle the challenging XFNL task.

% In addition, we conducted an experiment where no explicit instruction was provided; only the financial statement and the target numeral value were given during fine-tuning. This approach, however, also exhibited inferior performance (Table ~\ref{table:ablation_study} \quad w/o instruction). These outcomes demonstrate the challenges of this task when relying solely on the sentence content and numerical value, further highlighting the efficacy of our proposed \model{} framework.

\if{0}
% old table
\begin{table}[!thb]
  \centering
    \resizebox{\linewidth}{!}{
      \begin{tabular}{l|c|c}
        \toprule
        Model & Macro-F1 & Hits@1 \\
        \midrule

        FLAN-T5-XL with LoRA & 75.77 & 90.84 \\
        \quad w/ S-BERT-L12 as Tag Matcher & 73.31 & 89.45 \\
        \quad w/ S-BERT-L6 as Tag Matcher & 72.71 & 88.38\\
        \quad w/o instruction prompt & 65.05 & 77.28 \\
        \quad w/o tag metadata & 61.21 & 73.81 \\

        \bottomrule
      \end{tabular} 
      }
   \caption{Results for our ablation studies.}
   \label{table:ablation_study}
\end{table}
\fi

Next, we experiment with setting the XBRL tags, instead of their documentations, as the target for FLAN-T5-Large. In other words, we no longer use the tag metadata. 
Accordingly, the \textit{Tag Matcher} module now compares the embeddings of the generated and ground truth tags (and not documentations).
From the last row in Table~\ref{table:ablation_study}, we again observe a drop in scores with 19.8\% $\downarrow$ in Macro-F1 and 18.7\% $\downarrow$ in Hits@1 compared to our best results.
This confirms our hypothesis that the more elaborate tag documentations allow for a clearer distinction than the corresponding tags (possibly differing only in few words) while tackling the challenging extreme classification task.

\begin{table}[hbt!]
\resizebox{\linewidth}{!}{
    \begin{tabular}{p{0.39\linewidth}|p{0.3\linewidth}|p{0.39\linewidth}|p{0.25\linewidth}}
        \toprule
        Sentence & Ground truth tag & \axml \space Pipeline prediction & FLAN-T5-Large with LoRA \\
        \midrule
        \midrule
        As of December 31, 2019 and 2018, we had a cumulative translation loss, net of tax of \$ 617 million and \$ \textbf{\textcolor{blue}{466}} million, respectively. 
        & accumulated other comprehensive income loss foreign currency translation adjustment net of tax
        & accumulated other comprehensive income loss defined benefit pension and other postretirement plans net of tax \textcolor{red}{\XSolidBrush}
        & accumulated other comprehensive income loss foreign currency translation adjustment net of tax \textcolor{green}{\Checkmark}\\
        \midrule
        We also have \$\textbf{\textcolor{blue}{4.54}} billion of non - cancelable contractual commitments as of December 31, 2019 related to network infrastructure
        % and our data center operations.
        & contractual obligation 
        & unrecorded unconditional purchase obligation balance sheet amount \textcolor{red}{\XSolidBrush}
        & contractual obligation \textcolor{green}{\Checkmark}\\
        \midrule
        At April 24, 2020, plan participants had approximately \$ \textbf{\textcolor{blue}{14}} million withheld to purchase the company's ordinary shares 
        % at 85 percent of its market value on June 30, 2020
        % the last trading day before the end of the calendar quarter purchase period.
        & others
        & share based compensation arrangement by share based payment award number of shares available for grant \textcolor{red}{\XSolidBrush}
        & others \textcolor{green}{\Checkmark}\\
        \bottomrule
    \end{tabular}
    }
    \caption{Challenging examples where \axml{} predicts a wrong tag, while \bestmodel{} predicts correctly. %\axml{} struggles with challenging examples.
    }
    \label{table:base_comp}
\end{table}

%Near similar tags also makes the task more difficult for the \textit{Tag Matcher}.\\

 % We observe \model \space achieves best performance when utilizing the t5-xxl sentence encoder as tag matcher.

\begin{table}[hbt!]
\centering
\small
\begin{tabular}{p{1.0\linewidth}  p{0.1\linewidth}}
\hline
% \midrule
%\textbf{Instruction:} First, read the task description. There could be multiple numerals reported in a financial statement......
\textbf{Sentence:} Our Board of Directors declared quarterly dividends per share of \$\textbf{\textcolor{blue}{1.45}}, \$1.32 and \$1.15, which were paid in each of the four quarters of 2019, 2018, and 2017.\\

% \textbf{Question:} What is the tag associated with the numeral \textbf{\textcolor{blue}{1.45}}?\\

\textbf{GT Tag Doc:} aggregate \textcolor{green}{ dividends paid} during the period for each share of common stock outstanding. 

\textbf{GT Tag:} common stock dividends per share \textcolor{green}{cash paid}\\

\textbf{FLAN-T5-Large with LoRA (Generated, No Tag Matcher):} Aggregate \textcolor{red}{dividends declared} during the period for each share of common stock outstanding.  \\
% \textbf{FLAN-T5-XL with LoRA (with Tag Matcher):} aggregate dividends declared during the period for each share of common stock outstanding.\\
\textbf{Predicted Tag:} common stock dividends per share \textcolor{red}{declared}\\
\hline
\midrule

\textbf{Sentence:} In fiscal year 2016, Bard paid the Company \$\textbf{\textcolor{blue}{121}} million towards the settlement of 11,000 of these claims. \\

% \textbf{Question:} What is the tag associated with the numeral \textbf{\textcolor{blue}{121}}?\\

\textbf{GT Tag Doc:} amount awarded \textcolor{green}{from other party} in judgment or settlement of litigation. \\

\textbf{GT Tag:} litigation settlement amount awarded \textcolor{green}{from other party}\\
\textbf{FLAN-T5-Large with LoRA (Generated, No Tag Matcher):} Amount awarded \textcolor{red}{to other party} in judgment or settlement of litigation\\
% \textbf{FLAN-T5-XL with LoRA (with Tag Matcher):} amount awarded to other party in judgment or settlement of litigation.\\
\textbf{Predicted Tag:} litigation settlement amount awarded \textcolor{red}{to other party}\\
\hline
\bottomrule

\end{tabular}
% \caption{Examples to show that subtle differences between ground truth tag (GT tag) and its documentation (GT Tag Doc), and the generated text, can lead to a wrong final tag prediction. The few-word differences between Ground Truth (green) and predicted text (red) are highlighted.
% More examples in the supplementary section.}

\caption{Examples to show that subtle differences between ground truth tag (GT) and the predicted tag. The few-word differences between Ground Truth (green) and predicted text (red) are highlighted.
More such examples are in the Appendix A.3}
\label{table:tab_err}

\end{table}

%\subsection{Qualitative Analysis of predicted labels}
\noindent \textbf{Qualitative Analysis of predicted labels:}
Table~\ref{table:base_comp} shows the predictions made by our best model and the closest baseline (\axml{}) for a few challenging instances. 
Our proposed method classifies relevant numerals (first two examples) correctly and is able to find irrelevant numerals (the last example) and correctly tag that as \textit{`Others`}, whereas the baseline struggles for each cases.

\begin{figure}[h!]
\centering	
{
\includegraphics[height=4.5cm]{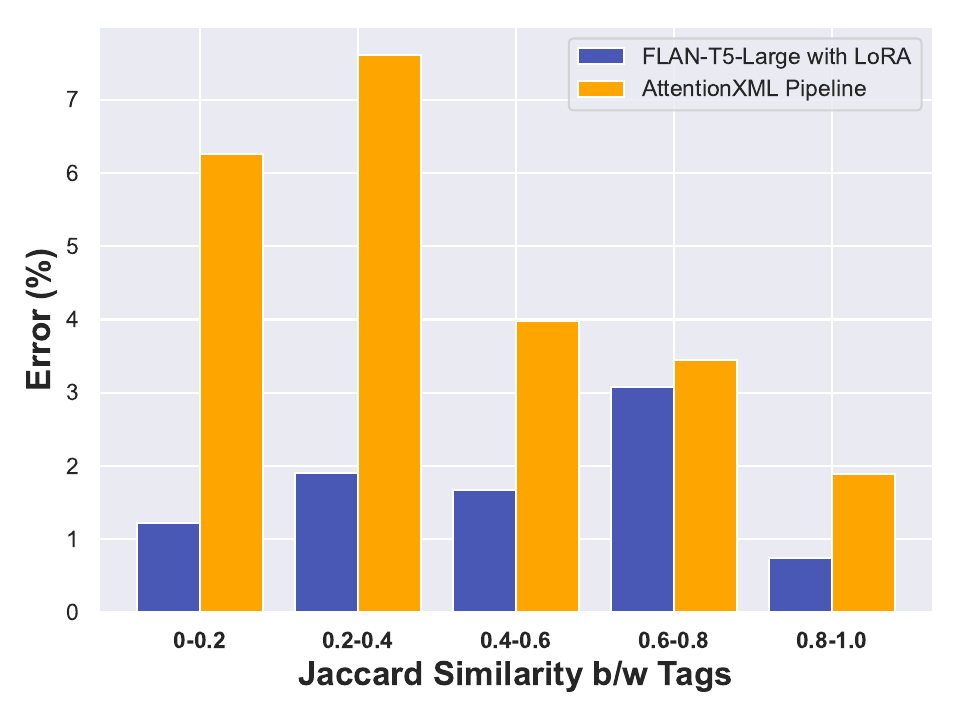}
}     
\caption{Comparing errors by best proposed model and those by the closest baseline. 
Even when our model generates incorrect tags, most of them are semantically very similar to the ground truth tags.
But \axml{} often generates completely unrelated tags.
%For a majority of errors by FLAN-FinXC, the wrongly predicted tag is very similar to the ground truth tag (Jaccard similarity between tag-words $\geq 0.6$). Whereas, most of the tags wrongly predicted by AttentionXML are very different from the ground truth tags (Jaccard similarity between tag-words $\leq 0.2$)
}
\label{fig:jaccard_error}
\end{figure}

\noindent \textbf{Error Analysis:}
%We conducted an analysis of the errors committed by FLAN-T5-XL with LoRA and the closest baseline.
To characterize the errors committed by our model, we ask -- \textit{when a model generates a wrong tag, how similar is the generated tag with the ground truth tag?} 
% That is, we distinguish between generating a tag similar to the ground truth and gross errors (generating a completely different tag). 
We quantify the similarity between a generated tag and the ground truth (GT) tag by the Jaccard similarity between the tag documentation words.
% (considering each tag documentation as a set/bag of words). 
Fig.~\ref{fig:jaccard_error} compares the errors by our best model and errors by the closest baseline. 
For a majority (60\%) of errors by our best model, the wrongly predicted tag is very similar to the ground truth tag (Jaccard similarity between tags $\geq 0.6$). 
%In fact, out of the 10\% inaccurate predictions by our model, in 6\% of cases, the Jaccard similarity between the ground truth and predicted tag documentation exceeds 0.6. 
Whereas, most of the tags wrongly predicted by AttentionXML are very different from the ground truth tags (Jaccard similarity between tags $\leq 0.4$).

%We find that even when our predictions are inaccurate (10\% of cases), these errors are often in close proximity. Referring to Fig.~\ref{fig:jaccard_error}, it becomes evident that in 6\% of cases, the Jaccard similarity score between the actual and predicted tag documentation exceeds 0.6, whereas for the error cases of the baseline models, the predicted tags are very distinct. 

% We also observe a marginal decrease in the performance of our best model when the target length exceeds 50 words. 

Finally, Table~\ref{table:tab_err} demonstrates instances where the predicted tag documentation closely resembled the ground truth (GT) tag documentation, but even minor variations led to a wrong final tag prediction. 
% In finance, subtle word changes can drastically alter context. 
As illustrated in Table~\ref{table:tab_err}, subtle differences between \textit{`from other party'} (in the GT tag, highlighted in green) and \textit{`to other party'} (in the prediction, highlighted in red), 
%or between \textit{`cash outflow'} and \textit{`cash inflow'},  
or between \textit{`dividends paid'} (in the GT tag, highlighted in green) and \textit{`dividends declared'} (in the prediction, highlighted in red) can lead to a wrong final tag prediction.  
To address these complexities, in future we plan to incorporate external financial knowledge.
% to enhance the model's performance.

\subsection{Comparison with ChatGPT}
\label{sec:appen:chatgpt}

We used ChatGPT (GPT-3.5 turbo) under a few-shot setting. For a fair comparison, we used the same prompt (shown in Fig.~\ref{fig:arch}) that is used to instruction tune our proposed models. The ChatGPT prompt structure was: \textit{<Instruction, Task statement, Question for a specific numeral>}.
In order to guide ChatGPT what to generate, we augment the prompt with 5 exemplars of (input, desired output) pairs, thereby making it a 5-shot setting.
Based on experimental observations, we set “desired outputs” to be the labels instead of label descriptions. Hence, in our final setting, no label descriptions were fed.
On the same 500 test set samples used to report ChatGPT’s performance (macro-f1 of 9.08) in Table~\ref{table:result_table}, we obtain macro-f1 score of \textbf{58.25} of our best model when trained to generate labels (note that Table~\ref{table:ablation_study} last row reports scores on the full test set). We believe that ChatGPT’s lower performance is owing to two main factors: 1)~It is not fine-tuned; rather, we use it in a 5-shot (few-shot) setting. 2)~ChatGPT was not specifically pretrained on financial data and the challenging nature of the FNXL dataset further makes it difficult.
Further analysis can be found in Appendix~\ref{sec:appen:chatgpt}.

% The challenging nature of the FNXL dataset further makes it difficult for ChatGPT to tackle the domain-specific task.
% Thus ChatGPT performs poorly on this complex task. 

\subsection{Experimental comparison among models} 
As stated previously, given the unavailability of generative baselines for the task, we had to first train and compare a suite of generative baselines as reported in Table~\ref{table:result_table}. This initial step allowed us to establish a fair comparison among models of similar sizes before showcasing the incremental performance gains with larger models. Table~\ref{table:exp_comp} presents a comparison of the number of trainable parameters in the different baselines, including the ones (generative) we trained, and our best-performing model. Also, a comparison of training times per epoch of the different models is shown in Table~\ref{table:exp_comp_trn_time}. T5-Base (220M), the weakest generative baseline trained by us, is comparable with Label Semantics, the largest existing non-generative baseline (218M). From Table~\ref{table:result_table}, T5-Base already outperforms Label Semantics with \textbf{18.56\%} Macro-F1 and \textbf{11.17\% }Hits@1 gains. T5-Base also outperforms AttentionXML (112M) substantially as reported above.
Also, while FLAN-T5-Large (780M) is larger than AttentionXML, our best results are obtained when we instruction-tune FLAN-T5-Large with LoRA, thereby drastically reducing the number of trainable parameters (0.59M), substantially lesser than AttentionXML (112M). Yet, we significantly outperform AttentionXML with\textbf{ 39.3\%} Macro-F1 and \textbf{17.2\%} hits@1 gains.

\begin{table}[hbt!]
\centering
\small 
\begin{tabular}{p{0.5\columnwidth}|c}
\toprule
Model & Trainable parameters \\ 
\midrule
FINER & ~109M \\
%\midrule
Label Semantics & ~219M \\
GalaXC & ~41M \\ 
%\midrule
AttentionXML & ~112M \\
%\midrule
 T5-Base & ~220M \\
 T5-Large & ~739M \\
 Flan-T5-Large & ~780M \\
\textbf{Flan-T5-Large w/ LoRA} & \textbf{~0.59M} \\
%\midrule
%payments to acquire held for sale real estate & 0.26 \\
\bottomrule
\end{tabular}
\caption{Comparison of number of trainable parameters between the baselines and \bestmodel{}}
\label{table:exp_comp}
\end{table}

\begin{table}[hbt!]
\centering
\small 
\begin{tabular}{p{0.45\columnwidth}|c}
\toprule
Model & Training time per epoch \\ 
\midrule
FINER & 12 hours \\
%\midrule
Label Semantics & 9 hours \\
GalaXC & 1.2 hours \\ 
%\midrule
AttentionXML & 4 hours \\
%\midrule
\textbf{Flan-T5-Large w/ LoRA} & \textbf{56 minutes} \\

\bottomrule
\end{tabular}
\caption{Comparison of training time per epoch between the baselines and \bestmodel{}} 
\label{table:exp_comp_trn_time}
\end{table}

%% file: Sections/Conclusion.tex
\section{Conclusion}
This work proposes a generative approach to solve the financial numeric labelling task.
We propose a novel \model{} framework, that makes use of parameter-efficient instruction tuning of LLMs for this extreme labelling task. While comparing with the state-of-the-art models and various competitive baselines that we devise for the task, we find that our best model, Flan-T5-Large with LoRA, achieves huge improvements, providing a Macro-F1 of 66.23\% as compared to the previously reported best numbers of 47.54\% for FNXL dataset. For adapting to other tasks, our approach requires only minor task-specific modifications to the input prompt, and the target output. Moreover, our approach offers flexibility in using either the class labels or label descriptions, depending on the available data. As potential future directions, we believe the scope to include more financial knowledge and integrate a human-AI feedback loop would be the way forward to improve performance of this challenging task.

% While the method can be applied more generally, the inclusion of label descriptions, when available, serves to enhance the overall performance, especially when there is significant textual/semantic overlap among the labels.
% }

% As potential future directions, we believe the scope to include more financial knowledge and integrate a human-AI feedback loop would be the way forward to improve performance of this challenging task.

% Though this work focuses on a particular problem in the financial domain, we believe our experiments bring out key insights into how generative LLMs can be used with parameter-efficient fine tuning approaches for challenging extreme classification / labelling problems in any domain, especially where the labels have associated semantics. 

% Various analyses and ablations have been reported to show the effectiveness of the proposed framework.
% Several analyses and ablations have demonstrated the efficacy of the proposed framework.

%% file: Sections/Limitations.tex
\section{Limitations}

%This work focuses on numerals found in SEC-mandated 10-K documents only. 
In this work, we have not integrated external financial knowledge to address the subtle differences between tags as identified in our error analysis. 
%Many companies tend to annotate text using their custom labels, which are absent in our dataset. 
Also, we have observed that labeling numerals solely based on sentence-level text (as done in this work) can be challenging, since the context depends on the surrounding paragraph, associated tables, and other elements which are not used in this work. Incorporating such elements as well as external financial domain knowledge into a financial numeral labeling model would be interesting future works.

%% file: Sections/Appendix.tex
\section{Appendix}

In this section, we provide some supplementary materials that enhance the content presented in the main paper. 
%We believe that the inclusion of these supplementary materials will facilitate a more comprehensive grasp of our research and findings.

\subsection{XBRL Tag Documentation}

Each XBRL tag is associated with a pre-defined textual description known as its `tag documentation'. Table~\ref{table:extra_tag_doc_example} shows some more  examples of XBRL tags and their documentations. Note that, while certain tag-pairs may exhibit subtle distinctions, their accompanying documentations vary significantly. Domain experts leverage the information provided in the documents during manual annotation. Our proposed model is also designed to utilize these documentations.

\begin{table}[hbt!]
\centering
\small
\resizebox{\linewidth}{!}{
\begin{tabular}{p{0.25\linewidth} | p{0.75\linewidth}}
%\begin{tabular}{c|c}
 \toprule
 Tag & Documentation \\
 \midrule
 \midrule
 common stocks shares issued & Total number of common shares of an entity that have been sold or granted to shareholders (includes common shares that were issued, repurchased and remain in the treasury). These shares represent capital invested by the firm's shareholders and owners, and may be all or only a portion of the number of shares authorized. Shares issued include shares outstanding and shares held in the treasury.\\ 
 \hline
 
common stock shares authorized & The maximum number of common shares permitted to be issued by an entity's charter and bylaws. \\ 
\hline
 \midrule
tax credit carry forward amount & The amount of the tax credit carryforward, before tax effects, available to reduce future taxable income under enacted tax laws. \\
\hline

tax credit carry forward valuation allowance & Amount of valuation allowance pertaining to the deferred tax asset representing potential future taxable deductions from tax credit carryforwards for which it is more likely than not that a tax benefit will not be realized. \\
\hline
\midrule
due to affiliate noncurrent	& Amount of receivables owed to an entity that is affiliated with the reporting entity by means of direct or indirect ownership, which are usually due after one year (or one business cycle, if longer).\\
\hline

due to affiliate current and non current & Amount of payable due to an entity that is affiliated with the reporting entity by means of direct or indirect ownership.\\
\hline
\midrule
notes payable related parties classified current & The amount for notes payable (written promise to pay), due to related parties. Used to reflect the current portion of the liabilities (due within one year or within the normal operating cycle if longer).\\
\hline
notes payable related parties current and non current	& The amount for notes payable (written promise to pay), due to related parties.\\
\hline
\midrule
% income loss from continuing operations before income taxes domestic	& The portion of earnings or loss from continuing operations before income taxes that is attributable to domestic operations.\\
% \hline

% income loss from continuing operations before income taxes foreign	& the portion of earnings or loss from continuing operations before income taxes that is attributable to foreign operations, which is defined as income or loss generated from operations located outside the entity's country of domicile.\\
 \bottomrule
\end{tabular}
}
\caption{Examples of XBRL tag documentations. While some tag-pairs exhibit very subtle distinctions, their accompanying documentations vary significantly. Our model takes advantage of the distinctions between the tag documentations.}
\label{table:extra_tag_doc_example}
\end{table}

\subsection{Comparison with ChatGPT}
\label{sec:appen:chatgpt}

With the emergence of ChatGPT, we were curious to check its performance on this task using the same instruction prompt for a few samples. As can be seen from the examples in Table~\ref{table:chatgpt_out}, ChatGPT struggles to correctly identify the appropriate tags in the majority of cases, and in some cases, it misunderstands the context entirely. This indicates that ChatGPT has still very limited performance in financial numerical tagging. 

\begin{table*}
\small
\begin{tabular}{p{1.0\linewidth} p{0.5\linewidth}}
\hline

\textbf{Instruction:} First, read the task description. There could be multiple numerals reported ....
% in a financial statement. Each number may or may not be associated with a particular tag. \\ 
 
\textbf{Sentence:} Now read the following financial statement.
 
 At April24, 2020 the estimated fair value was \$\textbf{\textcolor{blue}{27.1}} billion compared to a principal value of\$24.5 billion. \\
 
\textbf{Question:} What is the tag associated with the numeral \textbf{\textcolor{blue}{27.1}}?\\

\textbf{GT Tag:} \textcolor{green}{long term debt} fair value  \\
\textbf{ChatGPT:} \textcolor{red}{estimated} fair value \\
\textbf{\model:} \textcolor{green}{long term debt} fair value\\

\hline
\midrule

\textbf{Sentence:} Ordinary shares - par value \$0.0001, \textbf{\textcolor{blue}{2.6}} billion shares authorized, 1,345,400,671 and 1,341,074,724 shares issued and outstanding, respectively \\
 
\textbf{Question:} What is the tag associated with the numeral \textbf{\textcolor{blue}{2.6}}?\\

\textbf{GT Tag:} \textcolor{green}{common stock} shares authorized\\
\textbf{ChatGPT tag:} \textcolor{red}{ordinary} shares authorized\\
\textbf{\model:} \textcolor{green}{common stock} shares authorized\\
\hline
\midrule
\textbf{Sentence:} Share Capital Medtronic plc is authorized to issue 2.6 billion Ordinary Shares, \$0.0001 par value; 40 thousand Euro Deferred Shares, €1.00 par value; 127.5 million Preferred Shares, \$0.20 par value; and \textbf{\textcolor{blue}{500}} thousand A Preferred Shares, \$1.00 par value.\\
 
\textbf{Question:} What is the tag associated with the numeral \textbf{\textcolor{blue}{500}}?\\

\textbf{GT Tag:} preferred stock shares authorized\\
\textbf{ChatGPT:} a preferred shares authorized\\
\textbf{\model:} preferred stock shares authorized

\\
\hline
\midrule

\textbf{Instruction:} ...... \textbf{Sentence:} At April26, 2019, \$764 million of rebates were classified as other accrued expenses and \$\textbf{\textcolor{blue}{432}} million of rebates were classified as a reduction of accounts receivable in the consolidated balance sheets.\\
 
\textbf{Question:} What is the tag associated with the numeral \textbf{\textcolor{blue}{432}}?\\

\textbf{GT Tag:} contract with customer refund liability \textcolor{green}{current}\\
\textbf{ChatGPT:} \textcolor{red}{rebates reduction of accounts receivable}\\
\textbf{\model:} contract with customer refund liability \\
\hline
\midrule
\textbf{Sentence:} As of December 31, 2018, we expect to receive total future rental income of \$\textbf{\textcolor{blue}{203}} million related to noncancelable subleases for abandoned facilities. \\
 
\textbf{Question:} What is the tag associated with the numeral \textbf{\textcolor{blue}{203}}?\\

\textbf{GT Tag:} \textcolor{green}{operating leases future minimum payments receivable}\\
\textbf{ChatGPT tag:} \textcolor{red}{future rental income from noncancelable subleases}\\
\textbf{\model:} \textcolor{green}{operating leases} \textcolor{red}{rent expense sublease rentals 1} \\
\hline
\bottomrule
\end{tabular}
\caption{Comparison between ChatGPT and our best \model{}  model variant's prediction}
\label{table:chatgpt_out}

\end{table*}

\subsection{Error Analysis}

Table~\ref{table:appen:tab_err} demonstrates instances where the generated tag documentation (by our best model) closely resembled the ground truth (GT) tag documentation, but even minor variations led to a wrong final tag prediction.  In the financial domain, subtle word changes can drastically alter the context.  
As illustrated in Table~\ref{table:tab_err}, subtle differences between \textit{`from other party'} (in the GT tag, highlighted in green) and \textit{`to other party'} (in the prediction, highlighted in red), 
  or between \textit{`dividends paid'} (in the GT tag, highlighted in green) vs \textit{`dividends declared'} (in the prediction, highlighted in red), or between \textit{`cash outflow'} and \textit{`cash inflow'}, \textit{`loss'} vs \textit{`damages'} can lead to a wrong final tag prediction.

\begin{table*}
\begin{tabular}{p{1.0\linewidth}  p{0.3\linewidth}}
\hline

\textbf{Instruction:} First, read the task description. There could be multiple numerals reported in a financial statement......\textbf{Sentence:} On January 15, 2020, the parties agreed to a settlement in principle to resolve the lawsuit, which will require a payment of \$\textbf{\textcolor{blue}{550}} million by us and is subject to approval by the court. \textbf{Question:} What is the tag associated with the numeral \textbf{\textcolor{blue}{550}}?\\

\textbf{GT Tag Doc:} amount \textcolor{green}{of loss} contingency liability. \textbf{GT Tag:} \textcolor{green}{loss contingency accrual at carrying value}\\

\textbf{Flan-Large Generated:} Amount \textcolor{red}{awarded damages} contingency liability. \\
\textbf{Flan-Large +Tag Matcher:} amount of damages awarded to the plaintiff in the legal matter.\\
\textbf{Pred Tag-Words:} \textcolor{green}{loss contingency} \textcolor{red}{damages awarded value} \\

\hline
\midrule

\textbf{Sentence:} In fiscal year 2016, Bard paid the Company \$\textbf{\textcolor{blue}{121}} million towards the settlement of 11,000 of these claims. \textbf{Question:} What is the tag associated with the numeral \textbf{\textcolor{blue}{121}}?\\

\textbf{GT Tag Doc:} amount awarded \textcolor{green}{from other} party in judgment or settlement of litigation. \textbf{GT Tag:} \textcolor{green}{litigation settlement amount awarded from other party}\\
\textbf{Flan-Large Generated:} Amount awarded \textcolor{red}{to other} party in judgment or settlement of litigation\\
\textbf{Flan-Large +Tag Matcher:} amount awarded to other party in judgment or settlement of litigation.\\
\textbf{Pred Tag-Words:} \textcolor{green}{litigation settlement amount awarded} \textcolor{red}{to other party}\\
\hline
\midrule
\textbf{Sentence:} During the year ended December 31, 2017, we issued and repaid an aggregate of \$\textbf{\textcolor{blue}{12.3}} billion of commercial paper and had a maximum outstanding balance of \$1.5 billion under our commercial paper program. \textbf{Question:} What is the tag associated with the numeral \textbf{\textcolor{blue}{12.3}}?\\

\textbf{GT Tag Doc:} the \textcolor{green}{cash outflow due to repaying amounts} borrowed by issuing commercial paper.\textbf{GT Tag:} \textcolor{green}{repayments of commercial paper}\\

\textbf{Flan-Large Generated:} The \textcolor{red}{cash inflow during to theing short} borrowed under issuing commercial paper. \\
\textbf{Flan-Large +Tag Matcher:} the cash inflow from borrowing by issuing commercial paper.\\
\textbf{Pred Tag-Words:} \textcolor{red}{proceeds from issuance} \textcolor{green}{of commercial paper}\\
\hline
\midrule

\textbf{Sentence:} Our Board of Directors declared quarterly dividends per share of \$\textbf{\textcolor{blue}{1.45}}, \$1.32 and \$1.15, which were paid in each of the four quarters of 2019, 2018, and 2017, respectively.
\textbf{Question:} What is the tag associated with the numeral \textbf{\textcolor{blue}{1.45}}?\\

\textbf{GT Tag Doc:} aggregate \textcolor{green}{ dividends paid} during the period for each share of common stock outstanding. \textbf{GT Tag:} \textcolor{green}{common stock dividends per share cash paid}\\

\textbf{Flan-Large Generated:} Aggregate \textcolor{red}{dividends declared} during the period for each share of common stock outstanding.  \\
\textbf{Flan-Large +Tag Matcher:} aggregate dividends declared during the period for each share of common stock outstanding.\\
\textbf{Pred Tag-Words:} \textcolor{green}{common stock dividends per share} \textcolor{red}{declared}\\
\hline
\midrule
\textbf{Instruction:} .....\textbf{Sentence:} As of December 31, 2020, Duong met the held-for-sale criteria and loan receivable balance of \$1.3 billion, net of CECL reserve of \$\textbf{\textcolor{blue}{32}} million was reclassified  \textbf{Question:} What is the tag associated with the numeral \textbf{\textcolor{blue}{32}}?\\

\textbf{GT Tag Doc:} amount of allowance for credit loss on \textcolor{green}{accounts} receivable. \textbf{GT Tag Words:} allowance for doubtful accounts receivable\\

\textbf{Flan-Large Generated:}  othersmount of allowance for credit loss on \textcolor{red}{financing} receivable, A  \\
\textbf{Flan-Large +Tag Matcher:} amount of allowance for credit loss on financing receivable, classified as noncurrent.\\
\textbf{Pred Tag-Words:} allowance for notes and loans receivable noncurrent\\
\hline
\midrule
\end{tabular}
\caption{Error cases. Examples to show the subtle difference between ground truth (shown in green color) and generated tag docs (shown in red color) and predicted tag.   }
\label{table:appen:tab_err}

\end{table*}